# Detection of Sensor-To-Sensor Variations Using Explainable AI

Sarah Seifi[a,b], Sebastian A. Schober[b,c], Cecilia Carbonelli[b], Lorenzo Servadei[a,b], Robert Wille[a,d]

[a] *Technical University of Munich, Arcisstraße 21, 80333 Munich, Germany*
[b] *Infineon Technologies AG, Am Campeon 1-15, 85579 Neubiberg, Germany*
[c] *Johannes Kepler University Linz, Altenbergerstraße 69, 4040 Linz, Austria*
[d] *Software Competence Center Hagenberg GmbH (SCCH), Softwarepark 32a, 4232 Hagenberg, Austria*
Email of corresponding author: sarah.seifi@tum.de

*Abstract*—**With the growing concern for air quality and its impact on human health, interest in environmental gas monitoring has increased. However, chemi-resistive gas sensing devices are plagued by issues of sensor reproducibility during manufacturing. This study proposes a novel approach for detecting sensor-to-sensor variations in sensing devices using the explainable AI (XAI) method of *SHapley Additive exPlanations (SHAP)*. This is achieved by identifying sensors that contribute the most to environmental gas concentration estimation via machine learning, and measuring the similarity of feature rankings between sensors to flag deviations or outliers. The methodology is tested using artificial and realistic Ozone concentration profiles to train a *Gated Recurrent Unit (GRU)* model. Two applications were explored in the study: the detection of wrong behaviors of sensors in the train dataset, and the detection of deviations in the test dataset. By training the *GRU* with the pruned train dataset, we could reduce computational costs while improving the model performance. Overall, the results show that our approach improves the understanding of sensor behavior, successfully detects sensor deviations down to 5-10% from the normal behavior, and leads to more efficient model preparation and calibration. Our method provides a novel solution for identifying deviating sensors, linking inconsistencies in hardware to sensor-to-sensor variations in the manufacturing process on an AI model-level.**

## I. Introduction

The popularity of chemi-resistive gas sensing devices has been on the rise due to their low-cost production and flexibility as interest in air quality and its effect on health and well-being has grown in recent years [1], [2]. One major issue regarding graphene-based chemi-resistive gas sensors however is that sensor reproducibility is difficult to achieve in the manufacturing process [3]. Sensor-to-sensor variations can occur during the pre-treatment process or stem from differences in the graphene coatings arising during fabrication [4]. Especially when embedding AI on to the consumer sensor and preparing the machine learning model in the lab, it is vital to retrieve data from similar sensors used for model training and to remove potential outliers. A trend in the literature could be identified, where researchers are not only trying to find anomalies in their data but also try to explain their underlying cause which is known as Anomaly Reasoning (AR) [5]. Using common Machine Learning (ML)-based anomaly detection techniques, e.g. Autoencoders or Isolation Forest, deviations are detected [6]–[8]. Subsequently, XAI methods, e.g. *SHAP* [9], *LIME* [10] or *IntegratedGradients* [11], are applied to evaluate feature importance [12], [13]. By utilizing explainable techniques, the black-box ML model's outcome becomes more transparent, thereby increasing the trust of end users and ensuring compliance with regulations [14]. Nguyen et al. presented a Variational Autoencoder-based framework for network anomaly detection and used gradient-based explanations for anomalies [15]. Another way of identifying patterns that indicate potential anomalies in gas turbine systems was based on aleatoric uncertainty indicator and the *CUmulative SUm (CUSUM)* changepoint detection and which are afterwards explained using *SHAP* [16]. The authors showed that XAI can be used to improve the performance of traditional anomaly detection methods by providing more interpretable and actionable insights into the underlying causes of anomalies.

Our approach aims to combine both: explaining the model behavior over time on a local level, which is an extension of the global *SHAP* analysis done in [17], while also finding sensor variations and anomalies in the data. In this paper, we present a method using SHAP for detecting deviations in sensors that are difficult to reproduce in manufacturing. As an exemplary use-case, an in-house multi-channel chemi-resistive gas sensor that monitors the air quality of multiple environmental gases by estimating the gas concentrations was studied [17]. Two specific applications are analyzed in this work. The first application is to detect deviating sensors in the train dataset. By removing these deviations and retraining the model only with sensors that meet expected performance requirements, the overall performance of the model is improved. The second application is to find deviations in the test dataset, similar to how they appear in the manufacturing of the gas sensor. Overall, our method allows for a better understanding of how the different sensors contribute to the model estimation and can be used to improve the performance of the model. Furthermore, the sensor quality in the manufacturing process can be ensured.

## II. Methodology and Experiments

In this section, we describe the setup that was used to simulate the sensor behavior measuring specific Ozone concentration profiles. Subsequently, the methodology of the XAI outlier detection is explained.

## A. Sensor Setup

The analyzed sensor is an in-house multi-channel chemi-resistive environmental gas sensor estimating gas concentrations in parts per billion (ppb). It consists of several distinct sensor arrays, all constructed with pristine graphene as the base material [18]. Each sensor field is functionalized with different additional materials, leading to varying adsorption behaviors when in contact with gas molecules. This results in different electrical conductance changes which are then used to classify the different gases and extract features for the gas concentration estimation. To prevent potential saturation of the graphene, the sensor is heated up to a high temperature in a periodic manner using sinusoidal waves, which accelerates the desorption of gas molecules on the surface [17]. In this work, only one gas type, namely Ozone, has been analyzed in order to cancel out the effect of cross-sensitivities between different environmental gases and ensure a reliable proof-of-concept. Additionally, the sensor simulation tool based on a stochastic model developed in [19] was used to simulate the sensor behavior. This provides the ground truth regarding the sensor characterization, e.g. sensor defects or real sensor-to-sensor variations, and was used to retrieve measurement data.

## B. Measurement and Data Setup

Multiple sensors are simulated being in a single measurement chamber to simultaneously measure gas concentrations. In addition, these sensors also increase the dataset that is used to train, validate and test the model. By concatenating the response of the multiple sensors, the model is able to learn from a wider range of data, resulting in a more robust and reliable performance.

Afterwards, following time and frequency domain features were extracted for each sensor array: the relative Resistance (R), their first-order derivatives, i.e. the SLope (SL), the AMPlitude (AMP) and Phase Angle of the first harmonics (PA), and the Total Harmonic Distortion (THD).

In order to test our proposed sensor-variation detection method using XAI, two different types of concentration profiles were analyzed. Distinctly designed artificial profiles with high resolution, i.e. one concentration sample per minute, were used to characterize the sensor behavior. Here, the sensors were exposed to several Ozone profiles, known background, i.e. synthetic air, fixed humidity, temperature and pressure for the duration of 24 hours. Realistic Ozone concentration profiles of multiple cities were retrieved from the TOAR Database [20]. For each city, a typical spring Ozone concentration profile with the duration of 24 hours was used with a resolution of one Ozone concentration per hour. Out of the ten most prominent Ozone concentration shapes during a spring day, the first shape was chosen. One exemplary artificial and realistic Ozone concentration profile can be found in Fig. 1

Sensor-to-sensor variations means that even though sensors are manufactured in an identical fashion, they can still have variations in their resistive response to the same gas concentration profile. During the fabrication process, a variety of factors can contribute to this, e.g. the deposition of the sensing layer, the operation of the sensor heater or variations during the pre-treatment process [21], [22]. To simulate these variations throughout our investigations, the different parameters of the simulation model were altered in order change the sensitivity characteristics of the simulated sensors.

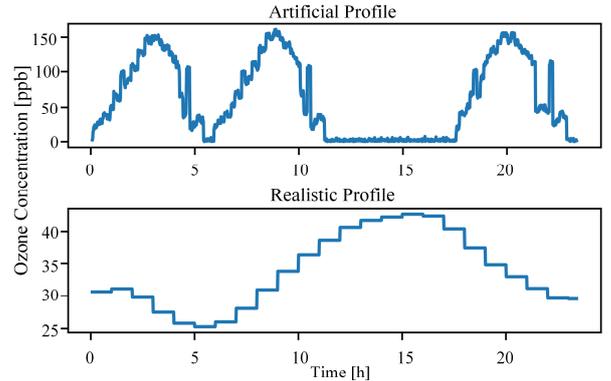

Fig. 1: Exemplary artificial and realistic Ozone concentration profiles used for model training.

## C. XAI-Sensor-Variation-Detection Setup

Due to the time-series nature of the data, a recurrent neural network, more precisely a *GRU* network, with one hidden layer and 40 units was used for the gas concentration estimation. After training the *GRU* and retrieving predictions for the gas concentrations, the *SHAP* algorithm was applied onto the model output.

The *SHAP* algorithm is a widely utilized model-agnostic method for explaining the output of ML models [9]. It uses the concept of Shapley values from cooperative game theory to determine the contribution of each feature to the model prediction. Shapley values are a method for fairly distributing a value among a group of individuals or features, based on their contribution to the overall outcome [23]. The algorithm calculates the average marginal contribution of each feature value across all possible coalitions which are combinations of features being present or absent. To do this, it first generates predictions for different coalitions with and without the analyzed feature, and then takes the difference between those predictions to calculate the marginal contribution of the feature. This process is repeated for all features, and the resulting values are the Shapley values which represent the importance of each feature on the model's prediction. This is defined as:

$$g(z') = \phi_0 + \sum_{j=1}^{M} \phi_j z_j' \quad (1)$$

with $g$ being the explanation model, $z' \in \{0,1\}^M$ is the coalition vector, $\phi_j \in \mathbb{R}$ is the Shapley value for a feature $j$, and $M$ is the maximum coalition size.

The *SHAP* algorithm provides explanations on a local as well as a global level which is achieved by averaging over all absolute local explanations:

$$I_j = \frac{1}{n}\sum_{i=1}^{n} |\phi_j^i| \quad (2)$$

with $I_j$ being the global explanation for feature $j$ and $n$ being the number of samples in the dataset. In this work, feature importance was analyzed on a local level over time and on a sensor level, respectively, and used to find sensor variations and defects. Local Shapley explanations ranking the input features according to their influence on the model predictions at each time step were calculated by training the *GradientExplainer* from *SHAP* which extends the *IntegratedGradients* method [11].

Since the *GRU* is trained with time-series data-batches, some datapoints are present in multiple batches and hence multiple local Shapley explanations for one datapoint were retrieved. These explanations were averaged and the resulting value was taken as the final local feature ranking $\phi_{j,avg}$:

$$\phi_{j,avg} = \frac{1}{m}\sum_{i=1}^{m}|\phi_j^i| \quad (3)$$

with *m* being the number of batches a specific datapoint occurs. As a next step, by summing over the absolute, averaged local Shapley values of each feature within one sensor and dividing by the number of samples of a sensor, a global feature ranking for each sensor $I_{j,sensor}$ was calculated:

$$I_{j,sensor} = \frac{1}{n_{sensor}}\sum_{i=1}^{n_{sensor}}|\phi_{j,avg}^i| \quad (4)$$

where $n_{sensor}$ represents the number of samples within one sensor. This global sensor feature ranking was then used to analyze the performance and behavior of one single sensor. Additionally, the ranking was utilized to compare the sensors' contribution to the model output and in this way detect variations and defects.

The comparison was done using three different similarity metrics, namely the *cosine similarity*, the *correlation coefficient* and the *Euclidean distance* between the global sensor feature rankings. Sensors with high deviations were then flagged as outliers and were removed. The *cosine similarity* is a measure of similarity between the non-zero vectors of an inner product space that measures the cosine of the angle between them and is defined as:

$$cosine\ similiarity = \frac{\sum_{i=1}^{n}x_i y_i}{\sqrt{\sum_{i=1}^{n}x_i^2}\sqrt{\sum_{i=1}^{n}y_i^2}} \quad (5)$$

with $x_i$ and $y_i$ being the values of dataset $x$ and y in position $i$, respectively.

The *correlation coefficient* quantifies the association between features of a dataset and is defined as:

$$correlation\ coefficient = \frac{\sum(x_i-\bar{x})(y_i-\bar{y})}{\sqrt{\sum(x_i-\bar{x})^2\sum(y_i-\bar{y})^2}} \quad (6)$$

where $\bar{x}$ and $\bar{y}$ are the mean values of dataset $x$ and $y$, respectively.

The *Euclidean distance* calculates the geometrical distance between two vectors and is defined as:

$$euclidean\ distance = \sqrt{\sum_{i=1}^{n}|x_i-y_i|^2}\ . \quad (7)$$

### D. Outlier Detection on Train and Test Dataset

Two applications were analyzed in this work. In the first application, variations were introduced into the train dataset, i.e. at sensor positions 2, 6, 8, 15 and 18. As can be seen in Fig. 2, using the train dataset with the included variations, the *GRU* network was trained and local Shapley explanations were retrieved. Afterwards, the global sensor feature rankings were compared using the similarity metrics and the dataset was pruned by removing outliers. XAI techniques have been utilized to prune networks, e.g. by performing model compression of the policy network of a reinforcement learning agent or by pruning *Convolutional Neural Networks (CNN)* using *LRP* or *DeepLift* [24]–[26]. The pruned dataset in this work was then used to train the *GRU* network and the performance of the network with and without the outliers were compared by evaluating it on multiple test gas concentration profiles. For this evaluation, the *Root Mean Squared Error (RMSE)*, the *Mean Absolute Error (MAE)* and the *Mean Absolute Percentage Error (MAPE)* was utilized. For the evaluation, the model was trained 6 times and the average scores were taken as a final result for comparison purposes.

For the second application, different degrees of variations were introduced into the test dataset simulating deviations occurring during fabrication. The outliers' sensor response deviates by 5% up to 30% compared to a normal sensor from lowest to highest disturbances at sensor positions 6, 2, 18, 15 and 8. The *GRU* network was trained with sensors meeting expected performance requirements and was tested on the test dataset including the defects. To test the methodology, various test datasets based on different gas profiles and defects were created.

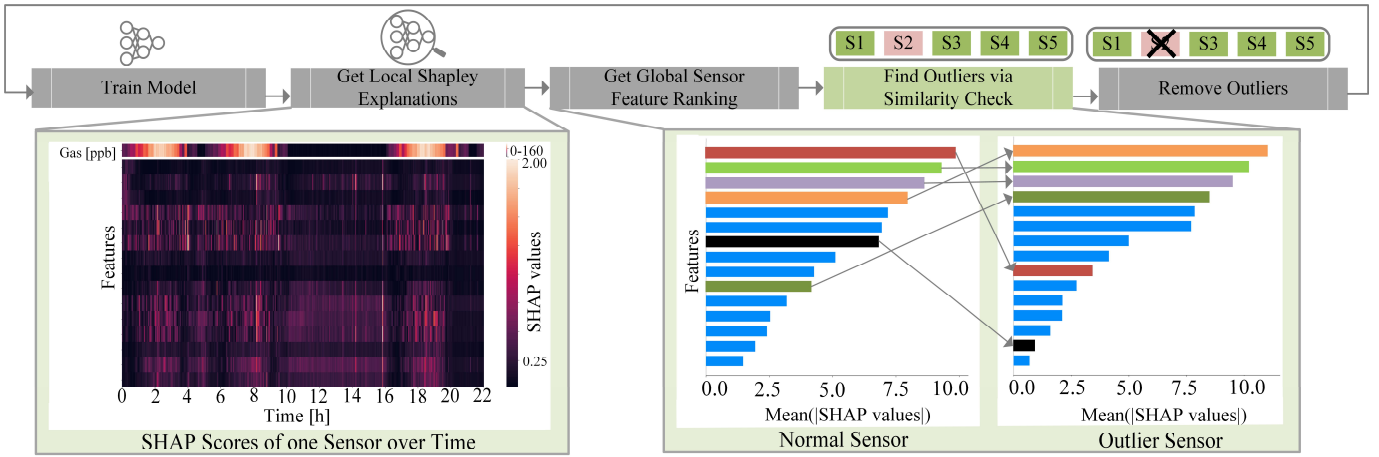

Fig. 2: Methodology of outlier detection and data pruning.

## III. RESULTS AND DISCUSSION

In this chapter, the results of the sensor-to-sensor variation detection using *SHAP* and dataset pruning is discussed. Two applications were studied: The first application is to detect deviating sensors in the train dataset and removing them. The second application is to find deviations in the test dataset.

### A. Outlier Detection in Train Dataset and Data Pruning

First, the sensor behavior using the artificially created Ozone profiles is discussed. Due to space limitations, only the results of one gas concentration profile is shown. In Fig. 3, the outcome of the *SHAP* algorithm of a normal and a sensor variation can be seen. From panel (a) it is observable that the THD is the most influential feature group regarding the model output followed by the phase angle. In the presence of gas, the sensor response is a shifted sinusoidal wave following the temperature curve of the sensor heater. This distortion can be captured using a *Fast Fourier Transform*. The THD feature finds the distortion percentage of the sensor's response signal from its fundamental wave shape. According to the global sensor feature ranking, this characteristic is fundamental and informative for correct gas predictions. Analyzing the *SHAP* heatmap in panel (b), as we call the heatmap of the local Shapley explanations over time, we can see that the slope feature is also important. It is not as influential over the entire course of the concentration profile (as could be seen in the global sensor ranking) but significant at specific events, i.e. fast changes in the concentration values. This information is lost in the global sensor feature ranking due to the averaging of the local Shapley explanations. Nevertheless, it validates the importance of the slope as a feature for the prediction.

In the case of deviating sensors, we can see in panel (c) that the THD feature group still has the most impact. This might lay in the fact that even though the sensor is not as responsive due to some manufacturing deviation or some fault in the calibration process, the THD feature as well as the amplitude are still able to capture the sensor dynamics in response to changing gas concentrations. Feature groups like the phase angle and the slope on the other hand lose on importance at a global sensor level. Nonetheless, from the *SHAP* heatmap it is evident that the slope feature is, as before, crucial at sharp concentration jumps and should hence not be omitted.

These results show, that normal and deviating sensors can be detected using the global sensor feature rankings, while the *SHAP* heatmaps provide a deeper understanding of the ML model prediction behavior. By comparing the global sensor feature ranking using the *Euclidean distance*, the *correlation coefficient* and the *cosine similarity*, deviating sensors with various degrees of variations were successfully identified. As an example, the results of the *Euclidean distance* are shown in Fig. 4. The *correlation coefficient* and especially the *cosine similarity* values between a normal and deviating sensor only slightly differ compared to the *Euclidean distance*. This can be seen as an indicator, that the *Euclidean distance* between two global sensor feature rankings is more conclusive. Nevertheless, even minor differences are sufficient to find the deviations in the train dataset.

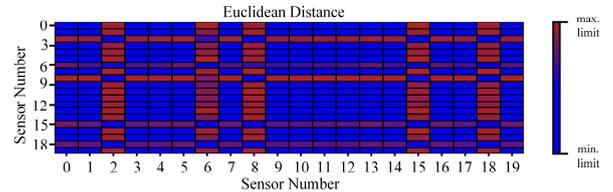

Fig. 4: Deviation detection results for an artificial train dataset using the *Euclidean distance*.

As a next step, the deviating sensors were pruned and the model was trained using the clean train dataset. In table 1, the averaged performance scores of six training runs for one exemplary concentration profile is shown. After removing the detected variations, the model trains for a shorter amount of time until convergence is reached while also achieving a better performance on multiple test datasets. It should be noted at this point, that even though the *GRU* was trained with highly varying sensors before the pruning process, it was still able to make good predictions for defect sensors, showing its good generalization abilities.

TABLE I. MODEL PERFORMANCE WITH AND WITHOUT OUTLIERS FOR AN ARTIFICIAL PROFILE

| Perform. Scores | rmse train [ppb] | mae train [ppb] | mape train [%] | rmse test [ppb] | mae test [ppb] | mape test [%] | duration train [s] |
|---|---|---|---|---|---|---|---|
| With outliers | 1.5 | 1 | 3.1 | 8.2 | 6 | 28.2 | 292.6 |
| Without outliers | 1 | 0.7 | 2.2 | 5.6 | 4.4 | 20.2 | 186.1 |

Similar results could be achieved when working with the realistic Ozone profiles from the TOAR database. To showcase this, the results of the Ozone concentration profiles of the cities Landen, Aarschot and Andenne in Belgium being used as train, validation and test dataset, respectively, can be seen in Fig. 5. As in the case of the artificial Ozone profiles, the THD and relative resistance features are the most impactful features on the model output. Feature groups like the slope barely impact the predictions since the realistic profiles only provide one sample per hour. Hence, sudden concentration jumps do not occur.

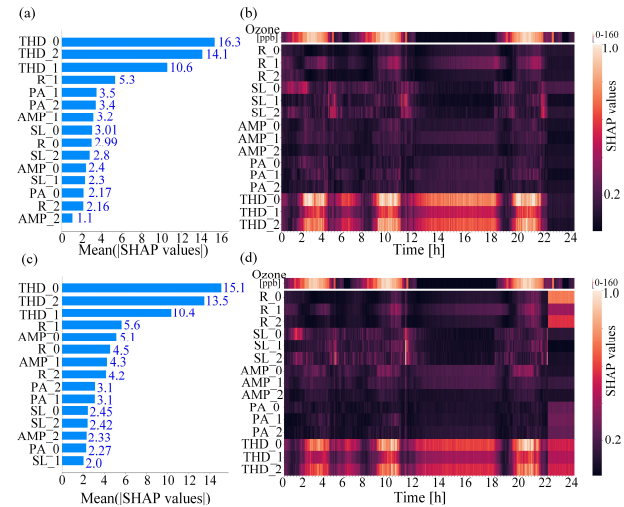

Fig. 3: *SHAP* feature importance results for an artificial profile: Global (a) and local (b) importances of a normal sensor and global (c) and local (d) importances of a deviating sensor.

For a sensor variation, the importance in the relative resistance and THD feature groups again drastically increases, as can be seen in panel (c). It should be noted that the predictions of the sensor deviations were not as accurate as those for the artificial profiles. Overestimation of low and underestimation of high concentrations occurs. Since the calculation of the Shapley values is based on the predictions that are made by the ML model and not the ground truth, the explanations retrieved in this case should be treated with caution. However, the THD again proves to be an impactful feature capturing the sensor dynamics well and leading to a good model performance.

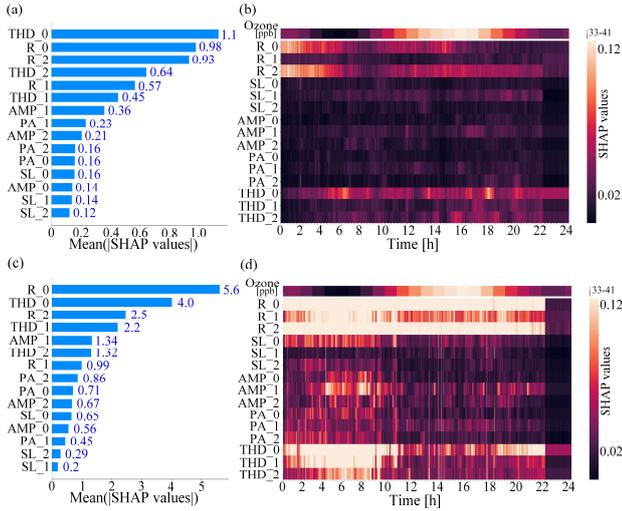

Fig. 5: *SHAP* feature importance results for a realistic profile: Global (a) and local (b) importances of a normal sensor and global (c) and local (d) importances of a deviating sensor.

In the realistic measurement setup, sensor variations were again correctly identified and afterwards pruned. The comparison of the averaged model performance using the unpruned and pruned train dataset can be found in table 2.

TABLE II. MODEL PERFORMANCE WITH AND WITHOUT OUTLIERS FOR A REALISTIC PROFILE

| Perform. Scores | rmse train [ppb] | mae train [ppb] | mape train [%] | rmse test [ppb] | mae test [ppb] | mape test [%] | duration train [s] |
|---|---|---|---|---|---|---|---|
| With outliers | 7.8 | 5 | 10 | 14 | 9.5 | 31.4 | 303 |
| Without outliers | 9.4 | 6.1 | 10.5 | 13 | 8.8 | 29.3 | 222 |

It is evident, that using the global sensor feature ranking and the *SHAP* heatmaps, we gain a more detailed insight into feature contributions towards the model output at specific time steps. Additionally, we correctly detect sensors with varying degrees of deviations enabling a subsequent hardware analysis. After removing detected deviations, we see an improvement in the model performance on various test datasets. This is achieved with less training samples leading to improvements in computational time and hardware resources.

## B. Outlier Detection in Test Dataset

This analysis was conducted for the artificial and the realistic concentration profiles, but due to space limitations, only the former is discussed. Nonetheless, similar results were also achieved for the realistic profiles. The results of the outlier detection method for one test dataset can be seen in Fig. 6. The value of the similarity measure matches the degree of the deviation. The higher the deviation, the higher is the *Euclidean distance* or smaller is the *cosine similarity* and the *correlation coefficient*. Using the *Euclidean distance*, disturbances higher than 10% can clearly be identified. To detect deviations at a 5% level, the sum of the *Euclidean distance* of one sensor to all others was calculated. This highlights the overall most deviating sensor and increases the detection sensitivity. However, the summed up *Euclidean distance* value of deviating sensor 6 and normal sensor 16 are very close to each other. This might be due to the stochasticity of the sensor simulation model which might lead to sensor 16 performing slightly different compared to the majority. By also adding the information of the *cosine similarity* and the *correlation coefficient*, sensor 16 can be ruled out as being an outlier. To increase the sensibility of the deviation detection in future work, comparing the differences in the local sensor Shapley explanations vectors instead of the global ones might be more informative. The *cosine similarity* and the *correlation coefficient* can also detect deviations higher than 10%, but the difference in the values of defect and normal sensors are very small and bring up the need of some lower thresholds. Overall, by including the information of all three similarity metrices, sensor defects down to a deviation of 5% can be identified.

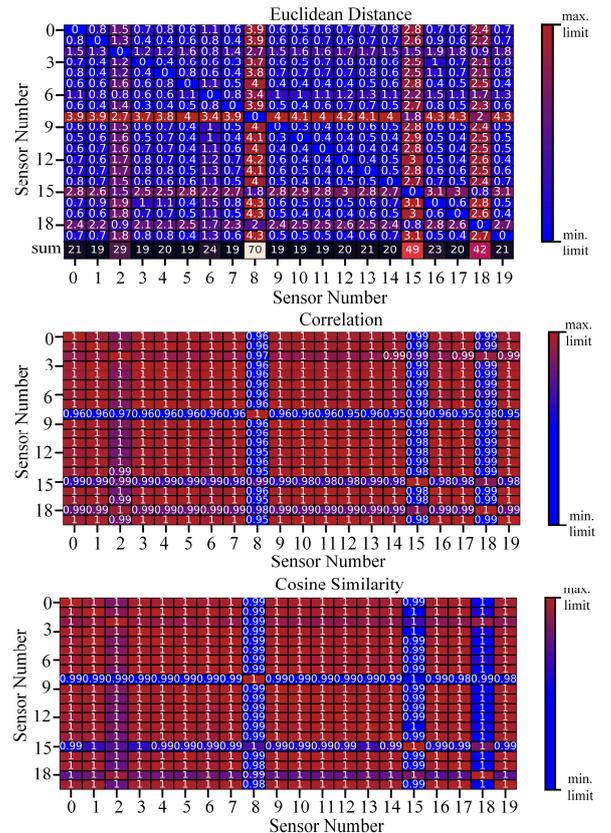

Fig. 6: Deviation detection results for an artificial test dataset: All similarity measures can detect sensors deviations inserted at sensor position 2, 6, 8, 15 and 18.

## IV. CONCLUSION

Our study demonstrates that Shapley explanations can be used to improve understanding of sensor and ML

model behavior by evaluating feature rankings on a sensor and local level at each time step. This approach helped highlight the importance of features like the slope, which may not appear significant globally, but are crucial at specific events such as sudden concentration jumps. Our results show that global sensor feature rankings are sufficient enough to not only better understand the sensor but also detect variations and defects. Additionally, we found that even though the *GRU* was intentionally trained with defect sensors, it was still able to perform well with a large train dataset, indicating its good generalization ability. Furthermore, by pruning the train dataset, we were able to achieve better model performance and save on hardware resources, which facilitates a faster model calibration in the lab. By finding sensor variations in the test dataset we could show that our method is suitable for detecting deviations in the manufacturing of difficult to reproduce sensors.

REFERENCES

[1] J. Saini, M. Dutta, and G. Marques, "A comprehensive review on indoor air quality monitoring systems for enhanced public health," *Sustain Environ Res*, vol. 30, no. 1, p. 6, Dec. 2020, doi: 10.1186/s42834-020-0047-y.

[2] A. D. Wilson and M. Baietto, "Applications and Advances in Electronic-Nose Technologies," *Sensors (Basel)*, vol. 9, no. 7, pp. 5099–5148, Jun. 2009, doi: 10.3390/s90705099.

[3] G. Lu et al., "Toward Practical Gas Sensing with Highly Reduced Graphene Oxide: A New Signal Processing Method To Circumvent Run-to-Run and Device-to-Device Variations," *ACS Publications*, Jan. 04, 2011. https://pubs.acs.org/doi/pdf/10.1021/nn102803q (accessed Feb. 06, 2023).

[4] S. A. Schober, Y. Bahri, C. Carbonelli, and R. Wille, "Neural Network Robustness Analysis Using Sensor Simulations for a Graphene-Based Semiconductor Gas Sensor," *Chemosensors*, vol. 10, no. 5, p. 152, Apr. 2022, doi: 10.3390/chemosensors10050152.

[5] P. Sharma et al., "Evaluating Tree Explanation Methods for Anomaly Reasoning: A Case Study of SHAP TreeExplainer and TreeInterpreter," in *Advances in Conceptual Modeling*, Cham, 2020, pp. 35–45. doi: 10.1007/978-3-030-65847-2_4.

[6] M. Carletti, C. Masiero, A. Beghi, and G. A. Susto, "Explainable Machine Learning in Industry 4.0: Evaluating Feature Importance in Anomaly Detection to Enable Root Cause Analysis," in *2019 IEEE International Conference on Systems, Man and Cybernetics (SMC)*, Oct. 2019, pp. 21–26. doi: 10.1109/SMC.2019.8913901.

[7] S. M. Tripathy, A. Chouhan, M. Dix, A. Kotriwala, B. Klöpper, and A. Prabhune, "Explaining Anomalies in Industrial Multivariate Time-series Data with the help of eXplainable AI," in *2022 IEEE International Conference on Big Data and Smart Computing (BigComp)*, Jan. 2022, pp. 226–233. doi: 10.1109/BigComp54360.2022.00051.

[8] L. Antwarg, R. M. Miller, B. Shapira, and L. Rokach, "Explaining Anomalies Detected by Autoencoders Using SHAP." arXiv, Jun. 30, 2020. Accessed: Dec. 19, 2022. [Online]. Available: http://arxiv.org/abs/1903.02407

[9] S. M. Lundberg and S.-I. Lee, "A Unified Approach to Interpreting Model Predictions," in *Advances in Neural Information Processing Systems*, 2017, vol. 30. Accessed: Jan. 17, 2023. [Online]. Available: https://proceedings.neurips.cc/paper/2017/hash/8a20a8621978632d76c43dfd28b67767-Abstract.html

[10] M. T. Ribeiro, S. Singh, and C. Guestrin, "'Why Should I Trust You?': Explaining the Predictions of Any Classifier." arXiv, Aug. 09, 2016. Accessed: Jan. 17, 2023. [Online]. Available: http://arxiv.org/abs/1602.04938

[11] M. Sundararajan, A. Taly, and Q. Yan, "Axiomatic Attribution for Deep Networks," in *Proceedings of the 34th International Conference on Machine Learning*, Jul. 2017, pp. 3319–3328. Accessed: Jan. 17, 2023. [Online]. Available: https://proceedings.mlr.press/v70/sundararajan17a.html

[12] L. Tronchin et al., "Explainable AI for Car Crash Detection using Multivariate Time Series," in *2021 IEEE 20th International Conference on Cognitive Informatics & Cognitive Computing (ICCI*CC)*, Oct. 2021, pp. 30–38. doi: 10.1109/ICCICC53683.2021.9811335.

[13] J. Sipple and A. Youssef, "A general-purpose method for applying Explainable AI for Anomaly Detection." arXiv, Jul. 23, 2022. Accessed: Dec. 19, 2022. [Online]. Available: http://arxiv.org/abs/2207.11564

[14] S. R. Islam, W. Eberle, S. K. Ghafoor, and M. Ahmed, "Explainable Artificial Intelligence Approaches: A Survey." arXiv, Jan. 23, 2021. Accessed: Jan. 22, 2023. [Online]. Available: http://arxiv.org/abs/2101.09429

[15] Q. P. Nguyen, K. W. Lim, D. M. Divakaran, K. H. Low, and M. C. Chan, "GEE: A Gradient-based Explainable Variational Autoencoder for Network Anomaly Detection," in *2019 IEEE Conference on Communications and Network Security (CNS)*, Jun. 2019, pp. 91–99. doi: 10.1109/CNS.2019.8802833.

[16] A. K. M. Nor, S. R. Pedapati, and M. Muhammad, "Application of Explainable AI (Xai) For Anomaly Detection and Prognostic of Gas Turbines with Uncertainty Quantification.," ENGINEERING, preprint, Sep. 2021. doi: 10.20944/preprints202109.0034.v1.

[17] S. Chakraborty, S. Mittermaier, C. Carbonelli, and L. Servadei, "Explainable AI for Gas Sensors," in *2022 IEEE Sensors*, Oct. 2022, pp. 1–4. doi: 10.1109/SENSORS52175.2022.9967180.

[18] A. Zöpfl, M.-M. Lemberger, M. König, G. Ruhl, F.-M. Matysik, and T. Hirsch, "Reduced graphene oxide and graphene composite materials for improved gas sensing at low temperature," *Faraday Discussions*, vol. 173, no. 0, pp. 403–414, 2014, doi: 10.1039/C4FD00086B.

[19] S. A. Schober, C. Carbonelli, A. Roth, A. Zoepfl, C. Travan, and R. Wille, "Toward a Stochastic Drift Simulation Model for Graphene-Based Gas Sensors," *IEEE Sensors Journal*, vol. 22, no. 12, pp. 11307–11316, Jun. 2022, doi: 10.1109/JSEN.2021.3114103.

[20] S. Schröder et al., "TOAR Data Infrastructure." https://b2share.fz-juelich.de/records/4d9a287dec0b42f1aa6d244de8f19eb3 (accessed Jan. 17, 2023).

[21] O. Tomic, H. Ulmer, and J.-E. Haugen, "Standardization methods for handling instrument related signal shift in gas-sensor array measurement data," *Analytica Chimica Acta*, vol. 472, no. 1–2, pp. 99–111, Nov. 2002, doi: 10.1016/S0003-2670(02)00936-4.

[22] M. Bruins, J. W. Gerritsen, W. W. J. van de Sande, A. van Belkum, and A. Bos, "Enabling a transferable calibration model for metal-oxide type electronic noses," *Sensors and Actuators B: Chemical*, vol. 188, pp. 1187–1195, Nov. 2013, doi: 10.1016/j.snb.2013.08.006.

[23] L. S. Shapley, "17. A Value for n-Person Games," in *17. A Value for n-Person Games*, Princeton University Press, 2016, pp. 307–318. doi: 10.1515/9781400881970-018.

[24] R. Xu, S. Luan, Z. Gu, Q. Zhao, and G. Chen, "LRP-based Policy Pruning and Distillation of Reinforcement Learning Agents for Embedded Systems," in *2022 IEEE 25th International Symposium On Real-Time Distributed Computing (ISORC)*, Västerås, Sweden, May 2022, pp. 1–8. doi: 10.1109/ISORC52572.2022.9812837.

[25] S.-K. Yeom et al., "Pruning by explaining: A novel criterion for deep neural network pruning," *Pattern Recognition*, vol. 115, p. 107899, Jul. 2021, doi: 10.1016/j.patcog.2021.107899.

[26] M. Sabih, F. Hannig, and J. Teich, "Utilizing Explainable AI for Quantization and Pruning of Deep Neural Networks." arXiv, Aug. 20, 2020. doi: 10.48550/arXiv.2008.09072.